\pdfoutput=1
\documentclass[11pt]{article}
\usepackage{authblk}
\usepackage{acl}

\usepackage{times}
\usepackage{latexsym}
\usepackage{url}
\usepackage{setspace}

\usepackage{hyperref}

\usepackage[T1]{fontenc}

\usepackage[utf8]{inputenc}

\usepackage{microtype}

\title{Interactivism in Spoken Dialogue Systems}
\author[*]{\textbf{Teresa Rodríguez Mu{\~n}oz}}
\author[*]{\textbf{Emily Ip}}
\author[*]{\textbf{Guanyu Huang}}
\author[*]{\textbf{Roger K. Moore}}
\affil[*]{Department of Computer Science, The University of Sheffield}
\affil[*]{\texttt{\{\href{mailto:trodriguezmunoz1@sheffield.ac.uk}{\texttt{trodriguezmunoz1}}, \href{mailto:eyjip1@sheffield.ac.uk}{\texttt{eyjip1}}, \href{mailto:ghuang10@sheffield.ac.uk}{\texttt{ghuang10}}, \href{mailto:r.k.moore@sheffield.ac.uk}{\texttt{r.k.moore}}\}@sheffield.ac.uk}}

\begin{document}
\maketitle

%% ABSTRACT --------------------------------------------------------------
\begin{abstract}
The interactivism model introduces a dynamic approach to language, communication and cognition. In this work, we explore this fundamental theory in the context of dialogue modelling for spoken dialogue systems (SDS). To extend such a theoretical framework, we present a set of design principles which adhere to central psycholinguistic and communication theories to achieve interactivism in SDS. From these, key ideas are linked to constitute the basis of our proposed design principles.
\end{abstract}

\noindent\textit{\textbf{Keywords}}: Spoken Dialogue System, interactivism, incremental dialogue, transactional model.

%% INTRODUCTION ----------------------------------------------------------
\section{Introduction}
In recent years, with the exponential growth of speech technologies such as Siri and Alexa, users have grown accustomed to the rigid dialogue schemes these devices offer. Thus, current human-robot interactions (HRI) are far from being conversational \cite{moore2016progress}. To optimise the effectiveness of HRI dialogues, researchers have worked on the accuracy of Automatic Speech Recognition (ASR), the naturalness of Text-to-Speech (TTS) modules and alternative dialogue frameworks. Incremental dialogue systems are one such alternative to attaining natural timing in conversation \cite{schlangen2011general}.

However, the quality of spoken interactions goes beyond increasing ASR accuracy and delivering timely responses. One must also have a better understanding of dialogue as a process, which is not linear, but rather, transactional \cite{pierce2009evolution}. Moreover, dialogue involves continuous, bi-directional interactions between the conversational agent, the contextual environment and the interlocutor via verbal and non-verbal signals \cite{moore2016introducing}. Hence, in this work, we suggest that high-performing SDS must embrace interactivism to demonstrate situational and social awareness.

%% BACKGROUND THEORIES --------------------------------------------------
\section{Background Theories}
\subsection{Interactivism: Dialogue as a Process}

Interactivism has favoured frameworks of process over models of substance \cite{bickhard2009interactivism}. This idea may translate to dialogue modelling, as dialogue is the process whereby ideas are exchanged among multiple social actors. It is also suggested that dialogue modelling is only one part of the interconnected modular system of an `intelligent' agent, which coexists in a given environment \cite{maturana1987tree}. Thus, the system behaves in an enactivist manner as a continuously and autonomously self-producing autopoietic entity \cite{moore2016introducing}. Therefore, future SDS design efforts should include \textit{autopoiesis} \cite{maturana1987tree} in the form of self-monitoring of the system's output (i.e., utterances produced by the agent) and its own current status. This has been attempted in incremental dialogue frameworks \cite{skantze2009incremental,schlangen2011general}, where self-monitoring feedback loops between the Contextualiser and Dialogue Modelling modules are employed to self-monitor, self-repair and monitor the user's speech production and non-verbal feedback signals.

Consequentially, the proposed interactivist framework presents implications for language and its use in human interaction. The framework is inherently social and interactional. Thus, the conversational agent needs to be imbued with the knowledge of these conventions to respond appropriately. Furthermore, future SDS cannot exclusively take in human responses in isolation; they must establish an awareness of the environment as it evolves through such an interaction and makes changes to itself accordingly \cite{moore2016introducing}.

\subsection{Transactional Model of Communication}
As an interactive process, spoken dialogue could be seen as transactional \cite{pierce2009evolution}. In comparison with linear and interactive dialogue models, the transactional model is the most dynamic. It considers dialogue as a cooperative process in which interlocutors exchange messages simultaneously. The dialogue is built upon shared experiences in culture, language and/or environment, allowing one to use less speech or even a single sound to achieve a successful interaction \cite{hawkins2003roles}. We propose that the transactional model may be the most preferred paradigm to achieve incremental dialogue. The conversational agent needs to be attentive and adaptive, which requires it to be able to adjust its language behaviours according to changes in the user and the environment. 

How can we make this adjustment happen? The inner workings of the brain reveal that thought is not linear; it is a process in which we constantly produce and shape ideas \cite{clark2014perceiving}. While the brain receives information from external resources, it tries to piece things together in a bottom-up way. It also attempts to guess incoming sensory data on the basis of what it knows about how the world is likely to be in a top-down manner. This Predictive Processing allows the agent to anticipate which actions to take, and to adjust its prediction upon its perception of embodied and environmental information \cite{clark2015surfing}. Hence, a predictive function should be employed to enrich SDS design. 

\subsection{Adjustment of Dialogue Behaviours}
In tandem with these \textit{intrapersonal} adjustments within the system, \textit{interpersonal} dynamics evolve as well. Entrainment describes how interlocutors become more similar to each other in their speech throughout a conversation \cite{levitan2013entrainment}, as speakers’ conversational behaviour tends to be influenced by that of the other interlocutor. For this reason, effort-based models \cite{lindblom1990explaining,moore2017toward} have been developed to account for humans' regulatory behaviour in everyday speech. These models consider what the speaker and the listener(s) share in common to adjust that effort. This closely relates to the Theory of Mind, which involves the ability to discern the mental states, including emotions, knowledge and beliefs of oneself and others \cite{woodruff1978does}. A conversational agent with such capabilities would be able to adjust to the user and build a rapport. This type of closed-loop dialogue system facilitates adaptive behaviours which are emergent, and not choreographed.

Furthermore, Gricean pragmatics can be considered to achieve more interactive behaviours in SDS. To more closely emulate human dialogue, Grice's maxims dictate that one should be as informative, concise, and relevant to the discussion as possible \cite{grice1975logic}. Moreover, ostensive communication (i.e., communication involving the expression and recognition of intentions via verbal or non-verbal cues) should be further explored in SDS design \cite{scott2017pragmatics}. This would allow the conversational agent to be aware of the multimodal protocols to open and close channels of communication and engagement.

% SUGGESTED DESIGN PRINCIPLES --------------------------------------------
\section{Design Principles for Transactional SDS}
Based on the theoretical background discussed, several design principles have been suggested below:
\begin{itemize}
    \setlength{\itemsep}{0.75pt}
    \item [(i)] \textbf{Conversational agents must have an incremental dialogue framework}: incremental SDS employ self-monitoring feedback loops to perform revisions on the system’s output (either covertly or overtly) and determine whether an utterance was spoken, interrupted by the user or revoked as a failed hypothesis.
    \item [(ii)] \textbf{Agents should adjust their communicative effort in dialogue}: if the conversational agent can identify the user’s abilities and adjust itself, then it could achieve autonomous, progressive learning of a user incrementally.
    \item [(iii)] \textbf{Conversational agents must be aware of the context of the conversation and have memory of past interactions with a user}. Short- and long-term information needs are essential in natural conversation.
    \item [(iv)] Further design of \textbf{multi-party SDS should consider ostensive behaviour} when engaging with different social actors. 
\end{itemize}

% CONCLUSION AND FUTURE WORK --------------------------------------------
\section{Conclusion and Future Work}

This work has briefly discussed interactivism and other central psycholinguistic and communication theories to improve the performance of current SDS. We have identified key design principles that the community may employ to design novel conversational agents, founded on the interactivist theoretical framework. Future work will involve the study of incremental dialogue systems and adapting them to align with the principles identified in this work.

% REFERENCES AND ACKNOWLEDGEMENTS ---------------------------------------
\section*{Acknowledgements}
This work was supported by the Centre for Doctoral Training in Speech and Language Technologies (SLT) and their Applications funded by UK Research and Innovation [EP/S023062/1]. Moreover, this work was partially supported by The Fulbright University of Sheffield Postgraduate Award.

\bibliographystyle{acl_natbib}
\bibliography{01_my_bibliography}

\end{document}